%% file: main.tex
\documentclass[sigconf]{acmart} 
\AtBeginDocument{%
  }
    
\usepackage[utf8]{inputenc}
\usepackage{amsmath}
\usepackage{tabularx}
\usepackage{booktabs}
\usepackage{array}
\usepackage{ragged2e}
\usepackage{hyperref}
\usepackage{xurl}
\usepackage{listings}
\usepackage{xcolor}
\usepackage{graphicx}
\usepackage{caption}
\usepackage{subcaption}
\usepackage{stfloats}

\hypersetup{
    colorlinks=true,
    linkcolor=blue,
    urlcolor=blue,
    breaklinks=true
}

\lstdefinestyle{llmprompt}{
  backgroundcolor=\color{lightgray!20}, 
  basicstyle=\ttfamily\footnotesize,    
  breaklines=true,                      
  frame=single,                         
  framerule=0.4pt,
  rulecolor=\color{gray},
  captionpos=b,                         
  keepspaces=true,                      
  showstringspaces=false,               
  columns=flexible,
}

\setcopyright{acmlicensed}
\copyrightyear{2025}
\acmYear{2025}
\acmDOI{XXXXXXX.XXXXXXX}
\acmConference[AI4DE@KDD '25]{International Workshop on AI for Data Engineering (AI4DE) co-located with KDD '25}{August 3, 2025}{Toronto, ON, Canada}

\begin{document}

\title{Leveraging LLMs to Create Content Corpora for Niche Domains}

\author{Franklin Zhang}
\affiliation{%
  \institution{Bellevue College}
  \city{Bellevue}
  \state{WA}
  \country{USA}}
\email{franklin.zhang@bellevuecollege.edu}

\author{Sonya Zhang}
\affiliation{%
  \institution{Eastside Preparatory School}
  \city{Kirkland}
  \state{WA}
  \country{USA}}
\email{szhang@eastsideprep.org}

\author{Alon Halevy}
\affiliation{%
  \institution{Google Cloud}
  \city{Mountain View}
  \state{CA}
  \country{USA}}
\email{alonhalevy@gmail.com}

\begin{abstract}
Constructing specialized content corpora from vast, unstructured web sources for domain-specific applications poses substantial data curation challenges. In this paper, we introduce a streamlined approach for generating high-quality, domain-specific corpora by efficiently acquiring, filtering, structuring, and cleaning web-based data. We showcase how Large Language Models (LLMs) can be leveraged to address complex data curation at scale, and propose a strategical framework incorporating LLM-enhanced techniques for structured content extraction and semantic deduplication. We validate our approach in the behavior education domain through its integration into {\bf 30 Day Me}, a habit formation application. Our data pipeline, named {\sc 30DayGen}, enabled the extraction and synthesis of 3,531 unique 30-day challenges from over 15K webpages. A user survey reports a satisfaction score of 4.3 out of 5, with 91\% of respondents indicating willingness to use the curated content for their habit-formation goals.

\end{abstract}





\maketitle

\input{sections/1_intro}

\input{sections/2_related}
\input{sections/3_overview}
\input{sections/4_gen}

\input{sections/5_search}
\input{sections/6_exp}
\input{sections/8_conclusion}

\section{Appendix}
\appendix
\input{sections/9_appendix}

\bibliographystyle{ACM-Reference-Format}
\bibliography{refs}

\end{document}

%% file: sections/1_intro.tex
\section{Introduction}

\noindent

Large Language Models (LLMs) have introduced a new level of intelligence to domain-specific applications, enabling systems to reason, generate, and interact in ways that were previously difficult or impossible. However, while LLMs provide the cognitive layer, domain-specific data and curated content continue to play a critical role, setting the direction, tone, and spirit of the application. The long-standing idea of treating the web as a vast corpus of knowledge~\cite{kilgarriff-grefenstette-2003-introduction, Gatto2014} has gained renewed relevance in this context. LLMs provide the potential to enhance this paradigm through intelligent, model-assisted data curation, making it possible to extract and organize high-quality, domain-relevant content at scale. 

Despite these advancements, several key challenges persist and call for solutions.
\begin{enumerate}
    \item How to source and locate a large volume of web content relevant to a domain? 
    \item How to extract data into a defined schema structure from largely unstructured and unorganized web content to enable an application?
    \item How to clean and deduplicate redundant data entries?
\end{enumerate}
In this paper, we demonstrate concrete solutions to these challenges through a particular application. We introduce a system that leverages LLMs for end-to-end data curation and application integration. Although designed for a specific domain, the architecture and methodology are extensible and can be adapted to a wide range of other domain-specific use cases.

\subsection{Example application}
\begin{figure*}
    \includegraphics[width=\textwidth]{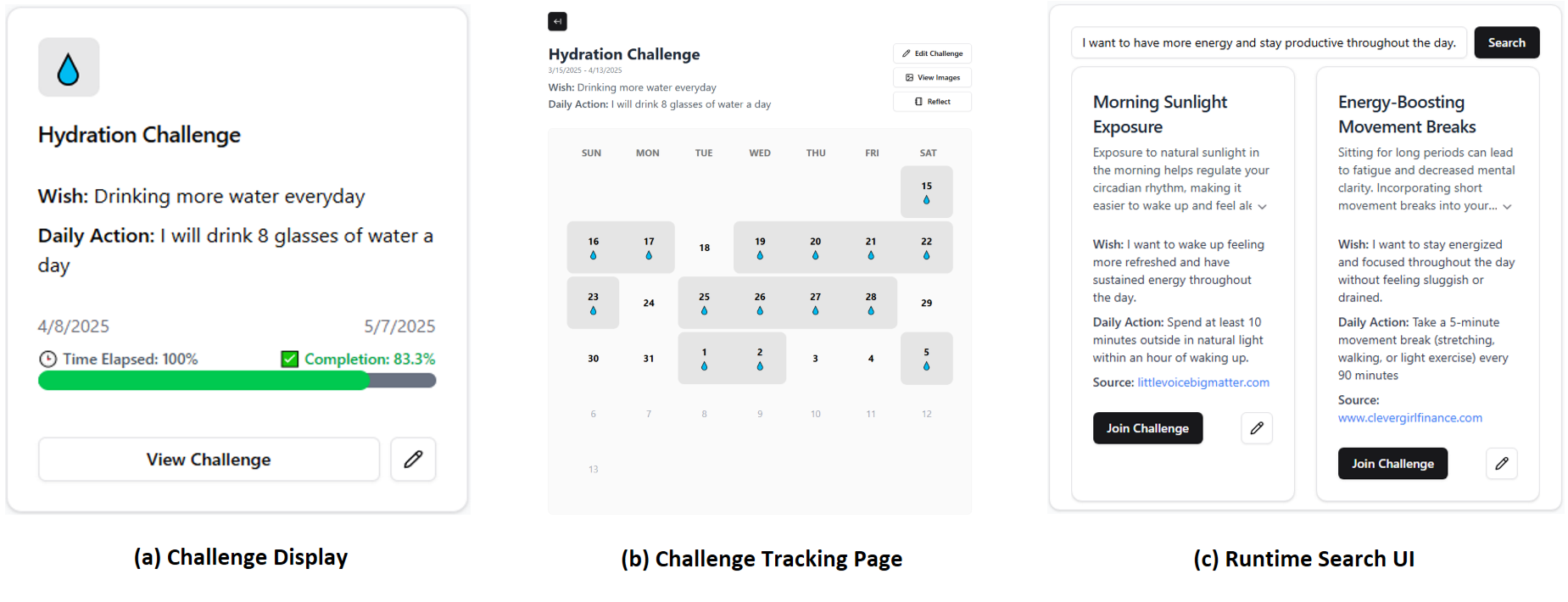}
    \caption{30 Day Me provides progress tracking (a) (b) and runtime challenge search (c).}
    \label{fig:crown-jewel}
    \Description[Screenshots of the 30 Day Me app showing challenge display, progress tracking, and challenge search features.]{Screenshots of the 30 Day Me app showing challenge display, progress tracking, and challenge search features.}
\end{figure*}

We developed {\bf 30 Day Me} (https://30day.me), a mobile app designed to help people achieve their long-term goals by turning them into manageable 30-day challenges. Each challenge begins with a personal goal, or “wish,” and pairs it with a simple daily action that consistently moves the user closer to that goal. The app helps users stay motivated and accountable by tracking their progress over time (see screenshot in Figure~\ref{fig:crown-jewel} (a) (b)). By focusing on daily, achievable steps, {\em 30 Day Me} promotes habit formation, skill development, and sustained personal growth through incremental progress.

A successful 30-day challenge hinges on having a clear and effective action plan that guides progress toward a meaningful goal. Research in goal-setting has shown that structured approaches, such as SMART goals,\footnote{\url{https://www.samhsa.gov/sites/default/files/nc-smart-goals-fact-sheet.pdf}} lead to better outcomes compared to vague or unstructured intentions. However, despite their benefits, crafting well-defined SMART goals can be difficult and time-consuming for many users.
 
\begin{figure*}
    \includegraphics[width=\textwidth]{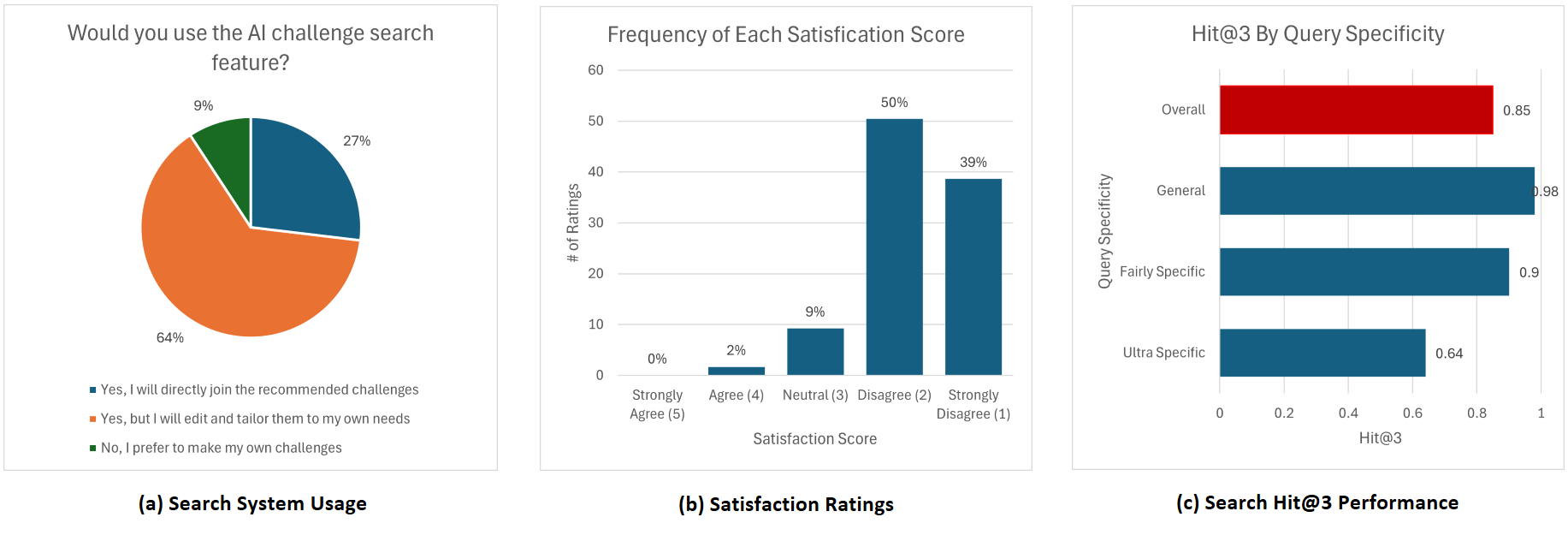}
    \caption{{\sc 30DayGen} search quality. User study shows that 91\% respondents would leverage the search results in creating 30-day challenges (a), and 89\% are satisfied with the search results (b). Offline evaluation exhibits overall hit@3=85\%.}
    \label{fig:crown-eval}
    \Description[Three charts: (a) Pie chart showing AI search feature usage intent, (b) Bar chart of user satisfaction scores, and (c) Bar chart illustrating search Hit@3 performance by query specificity.]{Three charts evaluating the 30DayGen search quality: (a) is a pie chart indicating that 64\% of users would directly use recommended challenges and 27\% would edit and tailor them, showing high usage intent. (b) is a bar chart displaying satisfaction ratings, with 50\% strongly agreeing and 39\% agreeing they are satisfied. (c) is a bar chart illustrating search Hit@3 performance by query specificity, with an overall hit rate of 0.85, and specific rates for general, fairly specific, and ultra-specific queries.}
\end{figure*}

Therefore, a core asset of the app is a corpus of 3,531 unique 30-day challenge ideas solicited from the web. 
At runtime, when a user submits a wish, the system searches this corpus and suggests challenges that are most likely to help users progress towards their goal (Figure~\ref{fig:crown-jewel} (c)). In a user survey, 91\% of the participants expressed that they would start from the search results to create challenges for their 30-day journey (Figure~\ref{fig:crown-eval} (a), details in Section~\ref{sec:survey}). 

\subsection{Corpus curation}
We present {\sc 30DayGen}, an automated end-to-end pipeline for domain-specific corpus curation. We started with posing 25 web search queries asking for a diverse assortment of challenges, and ultimately extracted and generated 3,531 unique challenges from online blogs and articles. In this process, we leverage LLMs in a novel way across all components of the system, including web filtering and collection, knowledge extraction, and semantic deduplication, transforming traditionally labor-intensive and error-prone tasks into tractable and reliable processes for better data collection and search.

First, initial web search with 25 queries related to 30-day habit forming resulted in 14,746 unique webpages, containing a significant amount of noise, where manual review can be tedious. We show that with representative examples, we can prompt an LLM to conduct few-shot learning for URL-level filtering to rapidly assess the likelihood that a page contains valid 30-day challenge ideas. This process effectively selected 953 promising webpages and blogs, and achieved a filter precision of 94\%.

Second, different webpages structure challenges in various ways and traditional information extraction methods~\cite{chang2006survey} can fall short. We show that a careful prompt tuning can invoke an LLM to extract structured information from original content following a defined schema. We extracted 11,792 challenges from the 953 pages (average 12 per page) and generated the wishes and daily actions for each challenge; our user survey shows a satisfaction of 4.5 (out of 5) regarding the clarity and understandability of the generated content.

Third, there is excessive content overlap across different web sources and it requires a nuanced understanding of action plans to identify duplicates. We propose an end-to-end semantic duplication algorithm, that leverages LLMs to rapidly identify mostly similar daily actions as duplicates. Specifically, we reduced the challenge set from 11,792 to 3,531 unique challenges (3.3x), achieving an F-measure of 0.890 in deduplication.

Finally, at runtime, the typical retrieval-then-ranking 2-step search
pipeline is inadequate to ensure that the returned challenges are perfectly aligned with the user's wish. We present a new search algorithm that invokes an LLM to semantically validate that top recommendations are not only relevant, but also helpful to the user's wish, adding a layer of intelligent filtering. An offline study shows a Hit@3 of 85\% and a NDCG of 0.80.

\subsection{Contributions}
To the best of our knowledge, this is the first paper that describes an end-to-end LLM-based data curation pipeline in building a corpus of user-facing content for specific domains. In particular, we make the following three contributions.
\begin{enumerate}
    \item We describe {\sc 30DayGen}, an LLM-driven pipeline with automated web data collection, structured information extraction, semantic deduplication, and goal-driven search. The pipeline enables constructing a corpus of 3,531 unique 30-day challenges solicited from 14,746 unique webpages in under two weeks.
    \item At the core of our data curation pipeline is a novel deduplication method that combines semantic embeddings with LLM-based judgment to suggest similar daily actions. Our deduplication method reduced the challenge set by 3.3x, achieving 0.890 F-measure in duplicate identification. 
    \item We demonstrate the critical role of the curated content corpus in the app {\bf 30 Day Me} for habit formation and goal achievement, which has both individual users and school users for educational purposes. A survey from 119 participants shows that 89\% of them are satisfied with the search feature, and 91\% of them will leverage the search results for challenge setting (Figure~\ref{fig:crown-eval}).
\end{enumerate}

Despite the system's focus on the habit formation application, we believe the prompt templates developed for key tasks, including webpage filtering, structured data extraction, and entry deduplication, are broadly applicable to other domains, especially niche domains with limited numbers of instances such as podcasts, online courses, self-help resources, and recipe collections.

%% file: sections/2_related.tex
\section{Related Work}
\paragraph{Automated data collection pipelines:}
The practice of automatically building specialized text corpora from the web is long-standing~\cite{kilgarriff-grefenstette-2003-introduction, Gatto2014}. Early tools like WebBootCat~\cite{baroni-2004, symseridou2018thewebasacorpus} demonstrate automated collection, albeit often requiring significant post-processing or relying on keyword limitations. Novel LLM capabilities have transformed the task, enhancing frameworks for web-scale data collection. For instance, Berkane et al.~\cite{berkane2025} establishes a human-in-the-loop framework to produce research-ready datasets according to user defined research topic. It leverages an LLM to generate relevant search queries, and utilizes a reranking model to evaluate relevance of retrieved web page title to the original query. As another example, Fei et al.~\cite{fei2025} presents a pipeline utilizing LLMs to crawl and collect relevant data by generating related questions, self-proposing answers and reasoning, to then create queries used to acquire more data. These two systems aim to serve downstream question answering or model training, and thus have a lower bar for the cleanness and uniqueness of the collected data. 
{\sc 30DayGen} distinguishes itself by focusing on curating user-facing content, which presents unique data cleaning challenges, as data need to be strictly relevant, structured, and semantically deduplicated.

\paragraph{Knowledge extraction:}
Once data is collected, a significant data editing challenge lies in transforming raw, often unstructured or semi-structured text content into a structured and usable format. While traditional Knowledge Extraction \cite{Weikum2010KGHarvesting} often targeted rigid (subject, predicate, object) triples~\cite{dong2023generationsknowledgegraphscrazy}, these approaches can be limited when dealing with the diverse and nuanced information prevalent in web data.
LLMs significantly enhance this process, enabling the integration of unstructured and diverse web data into a desired schema~\cite{wu2024learningextractstructuredentities, Dagdelen2024-co}. 
Our approach, {\sc 30DayGen}, applies LLM Knowledge Extraction capabilities to the specific domain of 30-day challenges to extract less rigidly defined concepts like a 'wish' and the associated 'daily action.' LLMs allow us to perform this extraction, plus generation and formatting based on the core challenge actions, effectively transforming unstructured web content into a structured challenge corpus.

\paragraph{Data integration:}
Another related area is Data Integration, combining data from different sources to form a unified corpus~\cite{Dong2015-bo, halevy2017}. Key issues addressed include schema heterogeneity through schema mapping, entity heterogeneity through entity linkage, and value heterogeneity through data fusion. Entity linkage is critical to {\sc 30DayGen}~\cite{Dong2015-bo}. The LLM enhances entity resolution using semantic comparisons, which is shown to significantly increase performance over traditional string matching methods \cite{peeters2024entitymatchingusinglarge}. We proposed a unique deduplication pipeline that relies on embedding similarity for initial matching and leverages LLMs to refine the results for difficult pairs.

%% file: sections/3_overview.tex
\section{Overview}
\label{sec:arch}

\begin{figure*}[t]
    \centering
    \includegraphics[width=0.95\linewidth, keepaspectratio]{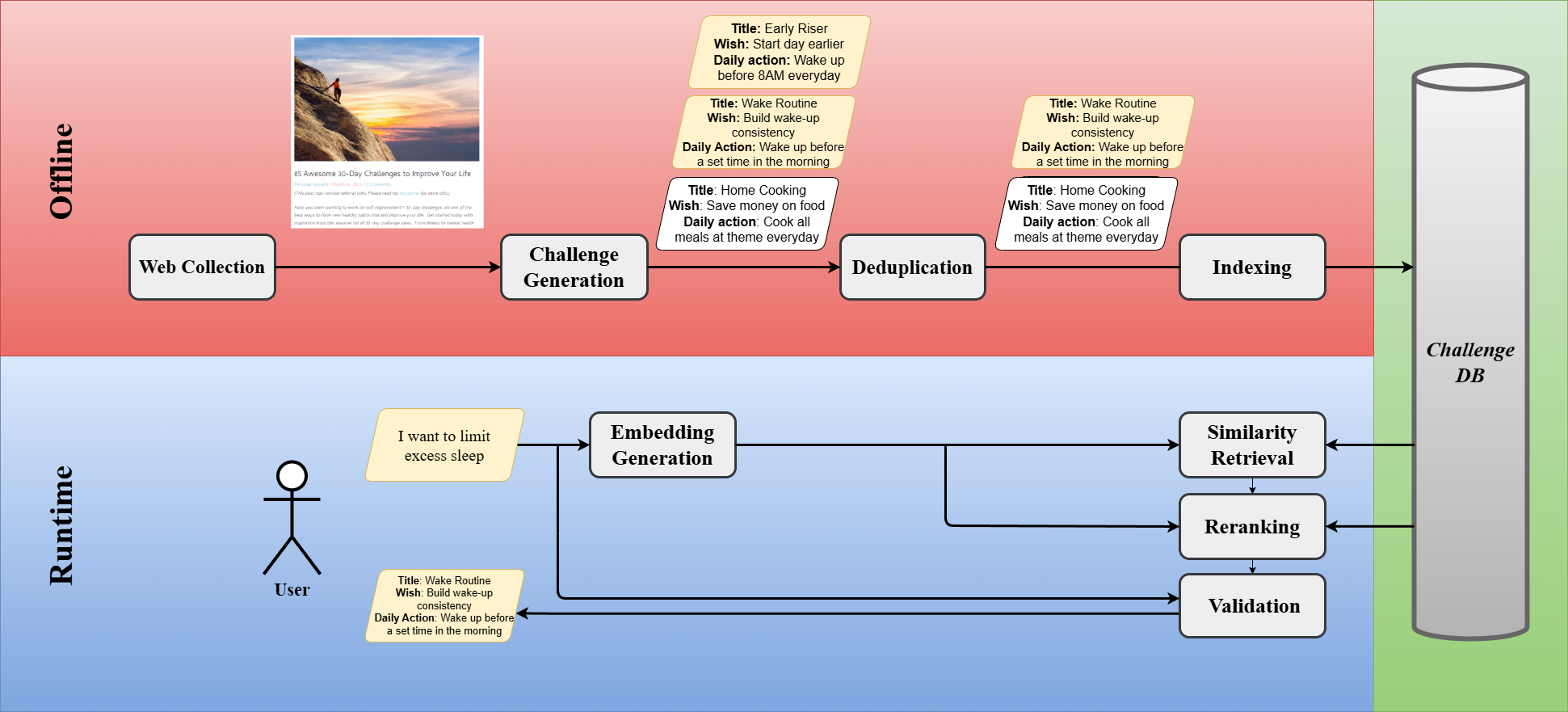}
    \caption{A high-level overview of the system architecture, illustrating the key components and their interactions.}
    \label{fig:arch}
    \Description[Architecture: Offline data processing and runtime user query pipelines for a challenge system.]{Diagram shows a challenge system. Offline pipeline: Web Collection, Challenge Generation, Deduplication, Indexing into Challenge DB. Runtime pipeline: User query, Embedding Generation, Similarity Retrieval from DB, Reranking, and Validation to show relevant challenges.}
\end{figure*}

\subsection{Problem definition}
We start by describing what is a 30-day challenge. A 30-day challenge is an action plan for a change that a user wishes to make to their lives everyday during a month-long period. It consists of two components: the {\em wish} and the {\em daily action}. The wish sets the goal that a user wishes to achieve (e.g., "feel less stressed"); the daily action suggests what one shall conduct to progress towards the goal (e.g., "meditate for 5 minutes daily").


We develop the {\sc 30DayGen} system that helps users find 30-day challenges relevant to their goals. The system takes a user query $Q$ describing their wish, searches a ChallengeDB corpus with web-sourced challenges, and outputs a list of challenges $\mathcal C$, best suited to the user wish.
Consider someone who struggles with low energy levels throughout the day and wishes to be more energetic. 
The top-2 suggestions are shown as examples in Figure~\ref{fig:crown-jewel} (c).

A critical step in building the {\sc 30DayGen} system is to populate ChallengeDB with high-quality challenges. We build ChallengeDB by extracting 30-day challenge ideas from resources available on the web. 


\subsection{{\sc 30DayGen} architecture}
We now describe the overall architecture of our system, depicted in Figure~\ref{fig:arch}. Our system has two parts: the {\em offline} system generates 30-day challenges by leveraging ideas from the web, and uses them to populate the {\em ChallengeDB}; the {\em runtime search} system takes the user's wish and searches the ChallengeDB to suggest  challenges.

The offline system is responsible for generating the challenge idea corpus. It has four components. First, the {\em Web Collection} component collects a set of web-pages containing concrete suggestions for 30-day challenges (For example, see \footnote{\url{https://www.sarahsteckler.com/blog/101-30-day-self-care-challenge-ideas}}). Second, the {\em Challenge Generation} component extracts ideas from the scraped pages, generating and formatting challenges in the required format. Third, the {\em Deduplication} component identifies and eliminates challenges deemed duplicates or too similar to another. Finally, the {\em Indexing} component populates the ChallengeDB with generated challenges and their semantic embedding representations.

The Runtime Search system takes a user query and returns a list of relevant 30-day challenges. It has 3 key components. First, the {\em Input Encoding} component creates a semantic representation of the user's input. Next, the {\em Challenge Retrieval} component matches the query embedding and challenge embeddings to identify relevant challenges. Finally, the {\em Ranking and Validation} component validates relevance of retrieved results and re-ranks them accordingly.

%% file: sections/4_gen.tex
\section{Offline Challenge Generation}
\label{offline}
In this section, we describe each component of the offline pipeline in detail: web collection, challenge generation, deduplication, and indexing.

\subsection{Web Collection}
The goal of the Web Collection component is to find a broad selection of blog and article URLs that suggest quality 30-day challenges. We look for web pages meeting the following criteria. 1) Pages should contain specific 30-day challenges rather than high-level promotion or discussion of the usefulness of habit formation. 2) Pages must provide challenges with repeatable, daily action items. 
3) Pages should offer challenges that represent a diverse assortment of potential user wishes and appeal to a wide range of interests.

Our approach to Web Collection consists of three steps: Composing search queries, collecting the web-pages from the searches, and filtering pages that fail our criteria.

\paragraph{Step 1. Search query composition.} 
We tested a variety of search queries suggested by GPT-4o and identified 25 unique ones that are effective for finding 30-day challenges: 
11 general and 14 tailored to specific themes. 
The full list of queries is included in Appendix~\ref{sec:search_query_list}.

\paragraph{Step 2. Search result collection:} 
We compiled search results 
using Bright Data's SERP API (Search Engine Results Page) to collect 25,000 web-pages (500 from each query), resulting in 14,746 unique results after deduplication.

\paragraph{Step 3. Webpage filtering:} 
The filter component takes each search result and decides whether it provides useful 30-day challenges following the criteria outlined above. 
We first remove blocked base domains including social media sites (e.g. YouTube, Pinterest) and popular e-commerce sites (e.g. Amazon, eBay), which are unsuitable for acquiring useful 30-day challenges (complete list of blocked base domains in Appendix~\ref{sec:blocked_base_domains}).
We then 
prompted Google's Gemini 2.0 Flash with in-context learning 
to determine a \emph{Likelihood Score} between 0–10 that signifies the probability that a webpage is useful. The component resulted with 953 URLs. Prompt in Appendix~\ref{sec:supplementary-materials}. 

\subsection{Challenge Idea Generation}
This component receives an input list of URLs and outputs a structured list of challenges parsed from the webpage content with generated wish and daily action fields. We prompt LLMs to conduct step-by-step challenge generation.

\paragraph{Step 1. Scraping and parsing:}
First, we used Puppeteer\footnote{\url{https://pptr.dev/}} to crawl and retrieve the HTML of the pages, then scraped and parsed each web-page to obtain the text content from the page. To minimize token inputs when extracting challenges, we removed header and footer elements and saved the rest of the text content.

\paragraph{Step 2. Challenge extraction:}
Next, 
we invoked Gemini 2.0 Flash to analyze each article, emphasizing two important tasks: 1) extract the exact text from article content to formulate challenge titles and descriptions; 2) create a 30-day challenge with the wish and daily action. (Prompt in Appendix~\ref{sec:supplementary-materials})

\subsection{Deduplication}
A major challenge with retrieving challenges from various web-pages and blogs is that many articles share overlapping ideas in their content. The deduplication component receives a set of challenges $\mathcal C$, identifies and removes duplicates, and outputs a list of deduplicated challenges ${\mathcal C}^u$ prioritizing ideas with better descriptions.

We begin by defining what constitutes a \textit{duplicate}. Challenges that suggest largely similar daily actions are considered duplicates. For example, \textit{"cooking a new meal every day"} and \textit{"trying a new recipe every day"} are duplicates because they both essentially propose cooking something new daily. We aim to optimize for well-balanced performance on two key metrics: 1) {\em Precision}, the percentage of challenges removed that are in fact duplicates. 2) {\em Recall}, the percentage of duplicates that are identified and removed.

In the literature~\cite{Christen2012-yu}, deduplication typically proceeds in three stages. First, {\em Blocking} groups items using fast, approximate similarity comparison to identify potential duplicates with high recall. Second, {\em Matching} analyzes every pair in the same block to detect duplicates. Finally, {\em Clustering} groups matched pairs into clusters, where each cluster represents a unique entity.

As the number of challenges is relatively small, and embedding-based similarity comparison is efficient at this scale, we bypass the traditional {\em Blocking} stage. 
Our approach focuses on refining the {\em Matching} stage, to go from synthetic similarity, to embedding-based similarity, to deep semantic understanding for duplicate identification.  First, we preliminarily filter our challenge idea dataset to remove obvious duplicates with high string similarity. Second, we utilize FAISS~\cite{johnson2019billion}, an indexing library that provides fast similarity search for high-dimensional vectors, to find duplicate-pair candidates based on embedding similarity. Third, we analyze moderate-confidence pairs using an LLM to accurately determine similarity. These three steps are effectively integrated into a comprehensive pairwise matching process.  Finally, we perform {\em correlation clustering} 
to finalize a deduplicated collection of 30-day challenges. As we show in experiments (Section~\ref{sec:dedup}), this progressive matching approach is scalable and effective.

\paragraph{Step 1: Preliminary filtering}
We first eliminate obvious duplicates with high string similarity. This step helps save computational costs in subsequent steps while using minimal resources. 
We focus on the title and daily action fields in this step because of their consistent format and significant role in our duplicate definition. 
We normalize titles and daily actions 
and remove stop-words before computing string similarities. 
We compile the stop-word list by using a combination of preexisting lists and extensive custom entries 
such as \textit{30day, challenge, day, improve}, etc.

\paragraph{Step 2: Pair-wise similarity computation}
Next, we compute pairwise similarity scores to identify potential near-duplicate challenge pairs. 
We first convert challenges into their vector representation. Since daily action is the primary factor in determining duplicates, we use OpenAI's text-embedding-3-large model with the daily action of each challenge as input to generate embeddings. 

We then find vector similarity for every possible pair within the challenge list. We use FAISS's IndexFlatL2, as its {\em brute-force} search approach provides the highest possible accuracy with sufficient efficiency for relatively small datasets. 

\paragraph{Step 3: LLM-based matching}
With a list of potential challenge idea duplicates, we sample pairs in each similarity range and manually examine the percentage of true duplicates. Accordingly, we determine a high and a low threshold, dividing the pairs into three segments: pairs above the high threshold are considered \textit{matches}, pairs below the low threshold are considered \textit{non-matches}, and pairs between the thresholds undergo LLM-based matching for further evaluation.

We prompted Google's Gemini 2.0 Flash model and its ability to understand the nuance of action plans to accurately determine whether a pair is a duplicate. (Prompt in Appendix~\ref{sec:supplementary-materials})


\paragraph{Step 4: Clustering}
Using the map of all high-accuracy pairs, we cluster similar challenges together to co-locate duplicates and create a deduplicated list of challenges.

Our intuition is that imperfect duplicate determination means that similarity is not transitive. For example, if A and B are a pair, and B and C are a pair, it is entirely possible for A and C to be insufficiently similar. Ideally, we would utilize \textit{correlation clustering} to create optimal groups given pairwise constraints. However, correlation clustering is an NP-complete problem~\cite{bansal2004}. We therefore employ a greedy algorithm as an approximation. Each challenge idea is treated as a node, evaluated sequentially to determine whether to join a cluster or to create its own individual cluster. This decision is made using the following criteria: if a challenge idea does not match at least half of the nodes in any existing cluster, it is allocated into a new cluster; otherwise, the node is inserted into the cluster with which it has the highest number of matches.

Finally, with duplicate challenges co-located within their own clusters, we choose one idea from each cluster to create a challenge list. For single-item clusters, that challenge idea is selected. For clusters with multiple ideas, we prioritize the challenge idea with the longest description, with the assumption that its detail makes it more specific and helpful.

\subsection{Indexing}
The Indexing component is responsible for inserting generated challenges into the ChallengeDB, providing access to the Runtime Search suite. We reuse the embeddings and the index generated for deduplication purpose, and upload challenges to the ChallengeDB.

%% file: sections/5_search.tex
\section{Runtime Search}
The runtime system enables users to discover challenges tailored to their personal goals. Given a user’s input wish, the system retrieves and ranks relevant 30-day challenges from the ChallengeDB. It does so by encoding the input query, performing a semantic similarity search against stored challenge embeddings, and then ranking and validating the retrieved results. In this section, we detail the key components of the runtime system.

\subsection{Challenge Retrieval} 
The Challenge Retrieval component finds challenges most semantically similar to a user wish. It takes a user query as input and outputs a list of potential challenge idea results. We do this in two steps: 

\begin{enumerate}
  \item \textbf{Input encoding:} We first encode the user's wish into a vector representation using OpenAI's text-embedding-3-large model.
  
  \item \textbf{Similarity search:} We use cosine-similarity to find top-k challenges most semantically similar with user input, querying ChallengeDB with the input embedding.
\end{enumerate}

\subsection{Rerank}
While cosine-similarity retrieval effectively identifies challenges semantically similar to a user's input, it lacks the precision needed for accurate relevance-based ranking.  To refine the results, the Rerank component leverages the bge-reranker-v2-m3 model via a Pinecone API endpoint. This model receives the daily action field of candidate challenges and returns a reranked list sorted by textual relevance to the user's query.

\subsection{Validation} 
The goal of the Validation component is to finalize and refine challenges recommended to the user. Determining best-fit challenges based on a user's wish requires a nuanced understanding of the effect of various lifestyle adjustments. By themselves, the Challenge Retrieval and Rerank components can sometimes provide irrelevant suggestions that are ineffective for helping a user accomplish their wish. 

We notice this occurring in two conditions:

\begin{enumerate}
\item \textbf{Insufficient data in ChallengeDB:} The Challenge Database lacks generated challenge idea data relevant to the user's expressed desire. For example, a user wishes to "prepare for the SAT." However, no challenges in the sources used to generate the ChallengeDB pertain to standardized testing improvement. Consequently, related but unhelpful challenges such as "Sit at the breakfast table" or "Make a wish everyday" are returned.

\item \textbf{Intentional contradiction despite thematic overlap:}  The challenge idea and the user's wish share a thematic overlap, but directly contradict each other in terms of intent. For example, a user wants to "wake up feeling refreshed in the morning." The challenge "Wake up 30 minutes earlier" might exhibit high vector proximity due to the shared theme of waking up. However, this challenge suffers in utility because waking up earlier will likely worsen fatigue, directly conflicting with the user's desire to feel refreshed.
\end{enumerate}

We thus utilize the Validation component, leveraging an LLM which can better assess user intent and accurately verify relevance of challenge idea suggestions. 

We again use Google's Gemini 2.0 Flash model to analyze the retrieved challenges and remove those which are irrelevant. (Prompt in Appendix~\ref{sec:supplementary-materials})

%% file: sections/6_exp.tex
\section{Benchmarks and Experiments Setup}

Our {\sc 30DayGen} system gathers challenge ideas through 25 search queries, resulting in 14,746 unique webpages. LLM URL filtering narrows this to 953 promising articles and blogs, which are then crawled to extract an initial 11,792 formatted challenge ideas. Finally, a four-step deduplication process refines this list in less than 15 minutes, generating a finalized set of 3,531 unique challenge ideas.

To evaluate the quality of our 30-day challenge corpus, as well as the end-to-end search performance, we conducted experiments to answer three questions:
\begin{itemize}
    \item \textbf{Q1}: Can we effectively leverage LLMs to identify duplicate challenges to build a high-quality challenge corpus?
    \item \textbf{Q2}: How well does 30DayGen retrieve relevant and helpful challenges based on user wish input?
    \item \textbf{Q3}: How do users perceive the quality and utility of the curated corpus?
\end{itemize}



\subsection{Challenge corpus construction (Q1)}
\label{sec:dedup}

\subsubsection {Evaluation Setup}
\paragraph{Metrics:}
We compute the precision and recall of our challenge deduplication component.

\begin{itemize}
    \item Precision: Precision computes the percentage of removed challenges that are indeed duplicates. We randomly sampled 100 removed challenges, identified the representative challenge that remained in the final corpus, and manually annotated if they are duplicates.

    \item Recall: Recall computes the percentage of duplicate challenges that are removed. We randomly sampled 100 challenges that remained after the deduplication process, and decided if it is a duplicate as follows: for each challenge, we found the top-5 similar challenges based on embedding similarity from the original challenge list; we then manually decided if any of these challenges is a duplicate of the examined challenge. Let $m$ be the percentage of challenges that have an unremoved duplicate. We compute recall as:
\[
\frac{prec \cdot removed\%}{prec \cdot removed\% + m \cdot (1-removed\%) }
\]
\end{itemize}

\paragraph{Methods:}
We compared our deduplication solution with the following methods. 
\begin{itemize}
\item Our solution: Vector pairwise matching, refinement with LLM matching for pairs where the similarities falling between 0.625 (low threshold) and 0.7 (high threshold), followed by correlation clustering.
\item Baseline 1: MinHash pairwise matching\footnote{https://pypi.org/project/datasketch/}, followed by transitive-closure clustering.
\item Baseline 2: Vector pairwise matching with threshold 0.7 (high threshold), followed by transitive-closure clustering.
\item Ablation 1: Remove LLM refinement step, instead only performing vector pairwise matching and correlation clustering.
\item Ablation 2: Replace correlation clustering with transitive-closure clustering.
\end{itemize}

\subsubsection{Deduplication Performance}
\begin{table}[t]
\small
\centering
\caption{Deduplication Performance. Our solution obtains the highest F1-Score.}
\label{tab:dedupe_results}
\begin{tabular}{l c c c}
\hline
\textbf{Method} & \textbf{Precision} & \textbf{Recall} & \textbf{F1-Score} \\ \hline
MinHash baseline & \textbf{0.970} & 0.688 & 0.805 \\
Vector-sim baseline  & 0.100 & \textbf{0.999} & 0.182 \\\hline
\textbf{Our Solution} & 0.930 & 0.853 & \textbf{0.890} \\
- LLM matching     & 0.740 & \textbf{0.685} & 0.711 \\
- Correlation Clus. & 0.040 & 0.999 & 0.077 \\ \hline
\end{tabular}
\vspace{-.1in}
\end{table}

Table~\ref{tab:dedupe_results} shows our experimental results for the Deduplication pipeline.
We have made the following observations.
\begin{enumerate}
    \item Our deduplication shows high effectiveness on our corpus, achieving a precision of 93\% and a recall of 85\% for removing duplicate challenges. It significantly outperforms baseline solutions (by 8\% and 71\%) and ablated solutions (by 18\% and 81\%), demonstrating the importance of each component in our deduplication solution.
    \item MinHash based on syntactic string matching achieves the highest precision (0.970), but is too conservative in removing duplicates, thus obtains a very low recall (0.688). 
    \item Transitive closure clustering, used in Vector-sim baseline and Ablation without correlation clustering suffers from poor precision. This confirms that similarity between challenges is not transitive and demonstrates importance of a more sophisticated clustering method. 
    \item The LLM-matching step significantly improves pairwise matching, improving precision by 19\% and recall by 17\%. 
\end{enumerate}

\subsubsection{URL filtering} 
Finally, we evaluate the accuracy of our LLM webpage filtering, which reduced an initial set of 14,746 unique pages to only 953 pages. We randomly sampled 100 removed webpages and manually checked if each page contains valid 30-day challenge ideas. Our filtering precision is as high as 94\%, showing the effectiveness of our LLM-based filtering.

\subsection{Challenge search (Q2)}

\subsubsection{Search Evaluation Setup}
\paragraph{Query set:} We prompted Gemini 2.5 Pro Exp to compose a list of 100 search queries, comprising of 30 general wishes like \textit{"I want to boost my energy levels,"} 40 fairly specific wishes like \textit{"I want to stay properly hydrated,"} and 30 ultra-specific wishes like \textit{"I want to be able to hold a plank for 2 minutes"} (full list is included in Appendix~\ref{sec:supplementary-materials}).

\paragraph{Ground truth answers:}
We queried Challenge DB with an embedded representation of each of the generated queries and compiled 50 of the most semantically similar challenges (majority of questions have fewer than 50 answers). We then manually reviewed these potential answers for each query, marking each result as correct (relevant and helpful) and incorrect, thereby forming a ground truth of highly relevant challenges.

\paragraph{Metrics:} We evaluate the ranking of our search results through five metrics.
(1) \textit{NDCG~\cite{jarvelin2002ndcg}} evaluates the relevance and ranking of the results, weighing correct results at the front of compiled answers more heavily than those at the end: \textit{DCG} measures the correctness of a document based on its position in the result list; NDCG is the ratio of DCG to the best possible DCG with perfect ranking, otherwise known as \textit{iDCG}. 
(2) Hit@3: Percentage of queries with at least 1 correct answer in its top 3 output challenges.
(3) Precision@K: Percentage of output challenges that are correct in the top-K search results. We measure Precision@3 and Precision@20.
(4) Recall@K: Percentage of correct results that are included in top-K search results, where the number is computed as min($k$, num\_of\_correct\_results). We measure Recall@3 and Recall@20.
(5) F1-measure@K: Harmonic mean between precision and recall. We compute it using the equation:
$
F_1 = \frac{2 \cdot \text{Precision} \cdot \text{Recall}}{\text{Precision} + \text{Recall}}
$

\paragraph{Methods:} We compare our search solution with an ablation without LLM filtering that decides usefulness of a challenge to achieve the user's wish.

\subsubsection{Evaluation results}
\begin{table}[t]
\small
\centering
\caption{Runtime Search Performance: Metrics with and without LLM Filtering}
\label{tab:search_results}
\begin{tabular}{l|r|rrr}
\hline
\textbf{Metric} & \textbf{Overall} & \textbf{General} & \textbf{Fairly} & \textbf{Ultra} \\
 &  &  & \textbf{Specific} & \textbf{Specific} \\ \hline
\textbf{Hit@3}      & \textbf{0.848} & \textbf{0.983} & \textbf{0.900} & \textbf{0.644}  \\
\ \ - Filtering     & 0.818 & 0.983 & 0.862 & 0.594 \\\hline
\textbf{Precision@3} & \textbf{0.770} & \textbf{0.977} & \textbf{0.866} & \textbf{0.433} \\
\ \ - Filtering & 0.740 & 0.977 & 0.833 & 0.377 \\ 
\textbf{Recall@3} & \textbf{0.778} & \textbf{0.977} & \textbf{0.858} & \textbf{0.472} \\
\ \ - Filtering  & 0.763 & 0.977 & 0.833 & 0.455 \\ 
\textbf{F-msr@3} & \textbf{0.774} & \textbf{0.977} & \textbf{0.862} & \textbf{0.452} \\ 
\ \ - Filtering & 0.751 & 0.977 & 0.833 & 0.412 \\ \hline 
\textbf{Precision@20} & \textbf{0.738} & \textbf{0.970} & \textbf{0.832} & \textbf{0.380}  \\
\ \ - Filtering   & 0.652 & 0.965 & 0.745 & 0.216 \\ 
\textbf{Recall@20} & 0.721 & 0.916 & 0.752 & 0.484  \\
\ \ - Filtering   & \textbf{0.821} & \textbf{0.965} & \textbf{0.826} & \textbf{0.671} \\ 
\textbf{F-msr@20} & \textbf{0.729} & 0.946 & \textbf{0.790} & \textbf{0.426} \\ 
\ \ - Filtering & 0.727 & \textbf{0.965} & 0.783 & 0.327 \\ \hline 
\textbf{NDCG}    & \textbf{0.797} & \textbf{0.970} & \textbf{0.852}  & \textbf{0.551}  \\
\ \ - Filtering & 0.774 & 0.966 & 0.814 & 0.530 \\ \hline
\end{tabular}
\end{table}

\begin{figure}
    \centering
    \includegraphics[width=0.8\linewidth]{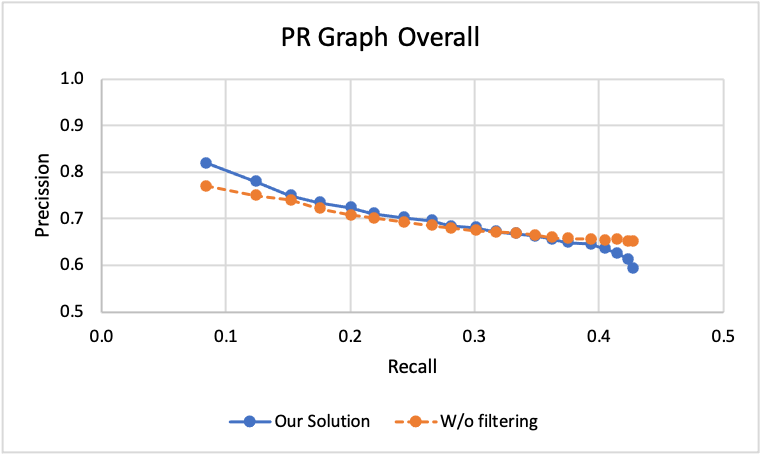}

    \centering 
    \includegraphics[width=0.8\linewidth]{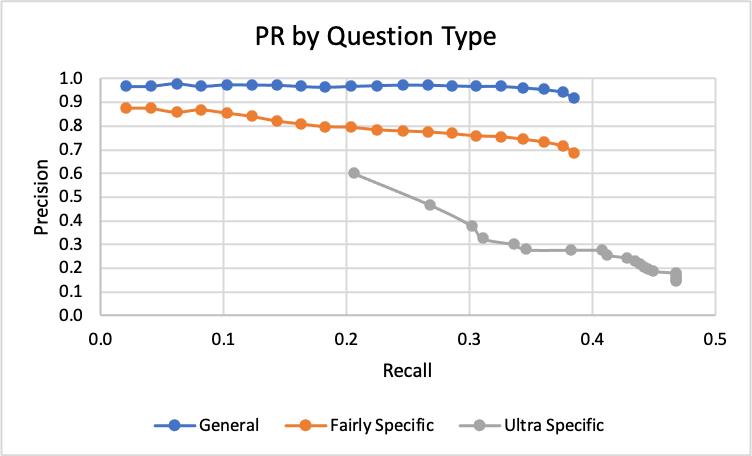}
    \caption{Precision-recall curves showing search performance with and without filtering and with different question specificity.} 
    \label{fig:pr-curves}
    \Description[Two precision-recall curve plots: the first shows overall search performance with/without filtering; the second details performance by question specificity.]{Figure displays two precision-recall graphs. Top: 'PR Graph Overall' compares 'Our Solution' (blue line) against 'W/o filtering' (orange line), with Precision vs. Recall. Bottom: 'PR by Question Type' shows PR curves for 'General' (blue), 'Fairly Specific' (orange), and 'Ultra Specific' (grey) questions.}
\end{figure}

Table~\ref{tab:search_results} shows our experimental results for Runtime Search system benchmarks and Figure~\ref{fig:pr-curves} shows the PR-curves of top-20 returned results. We have three observations:
\begin{enumerate}
    \item Our results demonstrate effective performance, with Hit@3 results of 0.848 overall, meaning for 85\% of questions we show relevant and helpful challenges in top-3 results, in particular high for general questions (0.983) and fairly specific questions (0.900). Additionally, we have high precision and recall for top-3 results, and even reasonable precision and recall for top-20 results.
    \item The model performance generally worsens as queries grow in specificity. We can see this with our NDCG values in particular, as the score drops from a high 0.970 to a medium 0.852 and to a mere 0.551, from general to fairly specific to ultra specific queries. This pattern is also reflected in the PR curves for different question types, as precision of ultra-specific queries rapidly decreases in comparison to both general and fairly-specific queries. This is due to the small number of challenges suitable for a narrow topic within our Challenge DB corpus. However, we still maintain a 0.644 Hit@3 score for ultra specific queries, showing that even for highly specific queries we still return helpful suggestions for two thirds of the questions. 
    \item Finally, we note that our solution utilizing LLM-filtering outperforms the ablated solution without it in all holistic metrics. The only regression is recall@20, where the LLM-filtering may be aggressive in filtering and hurts recall at higher K values. This observation is further reflected in the overall PR graph, as our filtered version delivers higher overall precision than the unfiltered version until towards the end. Still, our experimental results exhibit higher F-measure on more specific questions by removing unhelpful search results.
\end{enumerate}



\begin{figure}[t]
    \centering 

    \begin{subfigure}[t]{0.32\textwidth}
        \centering 
        \caption{User Study Feedback (Likert scale 1-5)} 
        \label{tab:user_study_results} 
        \small 
        \begin{tabularx}{\linewidth}{X|c} 
            \toprule
            \textbf{Question} & \textbf{Score} \\
            \midrule
            I'm satisfied with the challenge search system & $4.26$ \\ \hline
            The recommended challenges are clear and understandable & $4.49$ \\
            The recommended challenges are relevant to the searched goal & $4.45$ \\
            Following challenges recommended by the system would help achieve the searched goal & $4.18$ \\
            The search latency is acceptable & $4.46$ \\
            \bottomrule
        \end{tabularx}
    \end{subfigure}
    \hfill 
    
    \begin{subfigure}[t]{0.32\textwidth}
        \centering 
        \includegraphics[width=\linewidth]{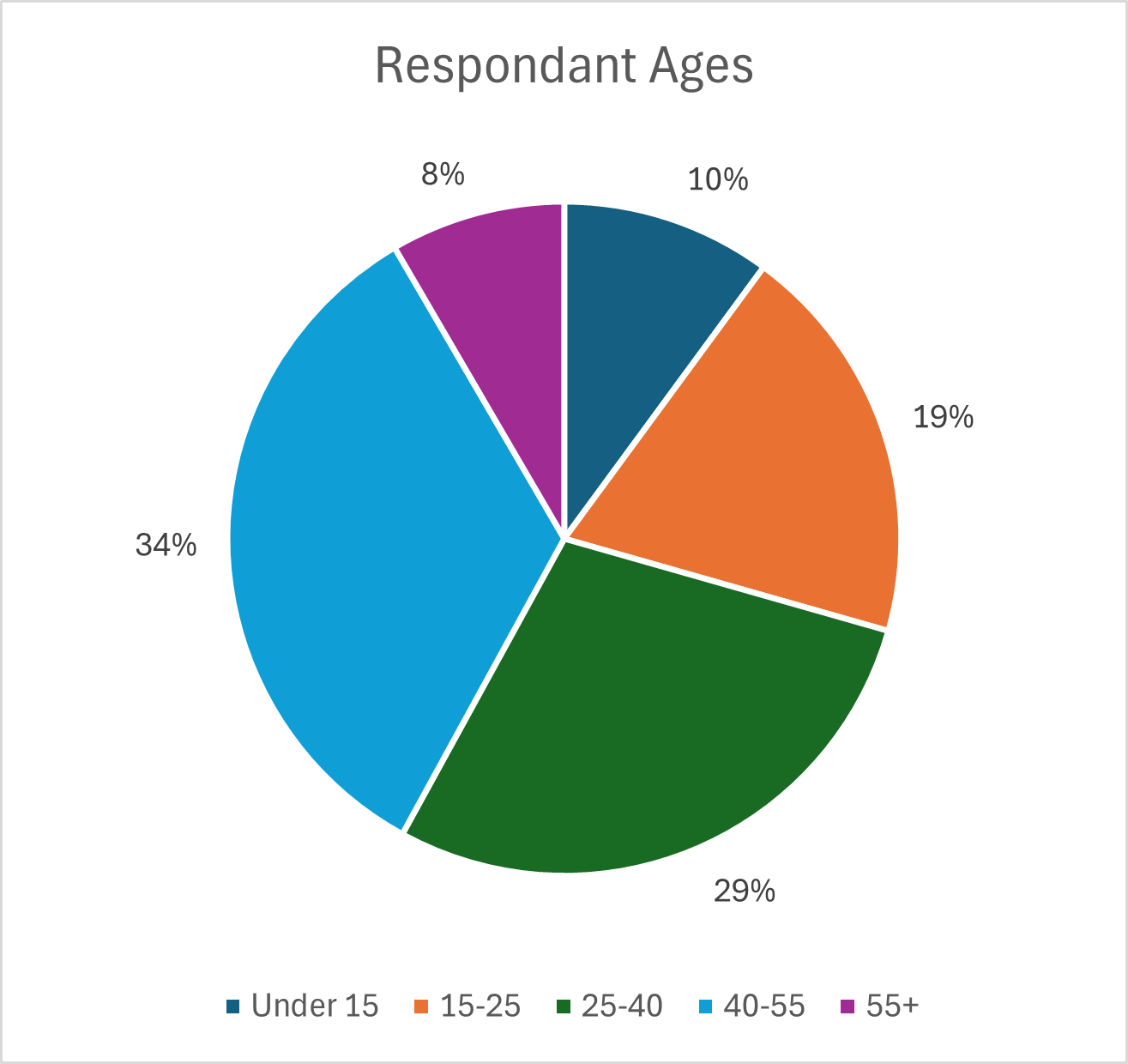} 
        \caption{Age distribution of survey respondents}
        \label{fig:age-pie} 
    \end{subfigure}
    \hfill 
    
    \begin{subfigure}[t]{0.32\textwidth}
        \centering 
        \includegraphics[width=\linewidth]{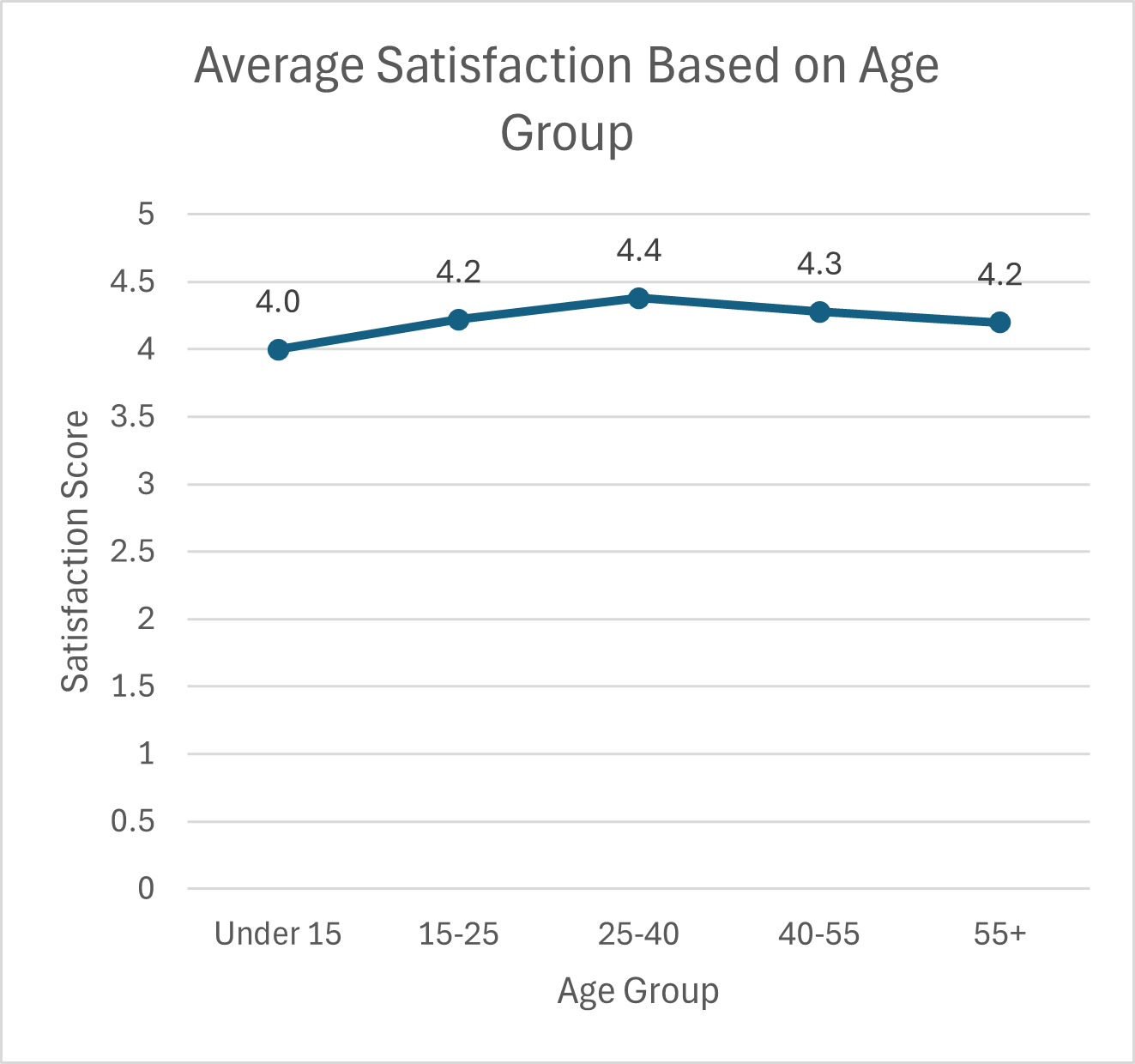} 
        \caption{Average satisfaction score respective to age group}
        \label{fig:satisfaction_score} 
    \end{subfigure}
    
    \caption{Overview of user study demographics, satisfaction, and feedback. Subfigure (a) shows user study feedback scores, (b) shows the age distribution of respondents, and (c) shows average satisfaction scores by age group.}
    \label{fig:user_study_summary} 
    
    \Description[User study overview: detailed feedback scores (table), respondent age distribution (pie chart), and satisfaction by age (line chart).]{This figure presents results from a user study across three subfigures. The first subfigure is a table titled 'User Study Feedback (Likert scale 1-5),' detailing user feedback scores on the challenge search system: 'I'm satisfied with the challenge search system' (4.26), 'The recommended challenges are clear and understandable' (4.49), 'The recommended challenges are relevant to the searched goal' (4.45), 'Following challenges recommended by the system would help achieve the searched goal' (4.18), and 'The search latency is acceptable' (4.46). The second subfigure is a pie chart titled 'Age distribution of survey respondents,' illustrating age demographics such as Under 15 (10\%), 15-25 (19\%), 25-40 (29\%), 40-55 (34\%), and 55+ (8\%). The third subfigure is a line chart titled 'Average satisfaction score respective to age group,' plotting satisfaction scores (Y-axis from 0 to 5, with observed averages typically between 4.0 and 4.4) against these age groups (X-axis).}
\end{figure}

\subsection{User Study on Search Experience}
\label{sec:survey}

To complement our quantitative search evaluation, we conducted a user study to evaluate user perception of the relevance and utility of search results. We distributed the survey\footnote{\url{https://www.30day.me/survey/30daygen}} on social media, particularly targeting student, parents, educational and fitness-oriented groups. We received 119 responses from a diverse assortment of age groups (see Figure~\ref{fig:age-pie}).

We set up our survey for respondents to experiment with the {\sc 30DayGen} search system. We then survey their impression of the system with three questions:

\begin{enumerate}
    \item We pose five statements regarding search results (see Table~\ref{tab:user_study_results}), and prompt them with a Likert scale table, asking for a value on a scale from 1 for least and 5 for most agreement.
    \item We ask "Would you use the AI challenge search feature?" to determine the utility of the search system, as shown in Figure~\ref{fig:crown-eval} (a).
    \item We provide an optional open response field to provide additional suggestions about the search system.
\end{enumerate}

\paragraph{User Study Evaluation Results:}

\begin{enumerate}
    \item Our results demonstrate that respondents are satisfied with the search system (4.3). In particular, they find search results are generally clear (4.5), relevant (4.5) and helpful (4.2). The satisfaction on the helpfulness of the answers (4.2) indicates high quality of corpus content and retrieval capability. They also find search latency generally acceptable (4.5), suggesting that LLM validation minimally affects response time. 
    \item User impressions suggest significant interest in using the challenge search system when creating a new challenge (only 9\% of respondents stated preference of starting from scratch). This showcases the utility of content corpus in general. 
    \item Open-ended comments provide insightful suggestions, such as {\em "make the search a conversational experience", "giving step by step instructions", "broken down into modular steps or blocks", "recommend a combination of daily actions to achieve the goal",} and {\em "more personalization"}. 
\end{enumerate}

%% file: sections/8_conclusion.tex
\section{Conclusion}
In this paper, we presented {\sc 30DayGen}, a novel solution that leverages LLMs for data acquisition, denoising and filtering, structured extraction from unstructured sources, and semantic deduplication to efficiently construct a specialized content corpus for habit formation. 
Our system successfully processed 14,746 webpages, harvested 3,531 unique, high-quality challenges, and provided search over the corpus with high quality (hit@3=85\%).
Our {\sc 30DayGen} system demonstrates how LLMs allow the end-to-end automation for large, structured data collection, cleaning, and refinement, enhancing the development of content-rich applications, and provides a methodological blueprint for creating similar data corpora. For future work, we plan to fully generalize our framework for curating content corpus to torso to tail domains.

%% file: sections/9_appendix.tex
\section{Supplementary Materials}
\label{sec:supplementary-materials}

All supplementary materials for this paper, including detailed LLM prompts, data lists such as search queries and blocked domains, and other supporting documentation, are available in our public GitHub repository.

The repository can be found at: \url{https://github.com/pigfyy/30DayGen-Supplementary-Materials}

\section{Search Query List}
\label{sec:search_query_list}

\subsection* {General Queries}
\begin{enumerate}
    \item fun and simple 30 day challenge ideas
    \item unique monthly challenge list for personal growth
    \item 30 day self improvement challenge ideas
    \item 30 day challenge ideas
    \item easy monthly challenges to try at home
    \item personal growth monthly challenge inspiration
    \item daily habit building 30 day challenge
    \item creative and productive monthly challenges
    \item motivational 30 day life improvement challenge
    \item list of fun challenges to do each month
    \item ideas for a different 30 day challenge each month
\end{enumerate}

\subsection*{Theme-Specific Queries}
\begin{enumerate}
    \item 30 day fitness challenge ideas
    \item monthly wellness challenge for healthy habits
    \item monthly learning challenge for self-education
    \item 30 day study challenge ideas for students
    \item monthly art challenge prompts for creativity
    \item 30 day writing challenge for creative practice
    \item monthly productivity challenge for better habits
    \item 30 day organization challenge for time management
    \item monthly sustainability challenge for eco-friendly living
    \item 30 day low waste lifestyle challenge
    \item monthly money saving challenge ideas
    \item 30 day no spend challenge for budgeting
    \item monthly kindness challenge for better relationships
    \item 30 day social skills improvement challenge
\end{enumerate}

\section{Blocked Base Domains}
\label{sec:blocked_base_domains}

\begin{itemize}
    \item YouTube
    \item Pinterest
    \item Facebook
    \item Instagram
    \item Amazon
    \item Reddit
    \item eBay
    \item LinkedIn
    \item Etsy
    \item Yelp
    \item TikTok
    \item Quora
\end{itemize}